\documentclass[lettersize,journal]{IEEEtran}
\usepackage{amsmath,amsfonts}
\usepackage{algorithmic}
\usepackage{array}
\usepackage[caption=false,font=scriptsize,labelfont=sf,textfont=sf]{subfig}
\usepackage{textcomp}
\usepackage{stfloats}
\usepackage{url}
\usepackage{verbatim}
\usepackage{graphicx}

\usepackage{multirow}
\usepackage{booktabs} 
\usepackage{makecell}
\usepackage{cite}

\def\BibTeX{{\rm B\kern-.05em{\sc i\kern-.025em b}\kern-.08em
    T\kern-.1667em\lower.7ex\hbox{E}\kern-.125emX}}
\usepackage{balance}
\begin{document}
\title{Fully Transformer Networks for Semantic Image Segmentation}
\author{ 
 Sitong Wu\textsuperscript{\rm 1, \rm 2 $*$}, 
 Tianyi Wu\textsuperscript{\rm 1, \rm 2 $*$},
 Fangjian Lin\textsuperscript{\rm 1, \rm 2, \rm 3 $*$},
 Shengwei Tian\textsuperscript{\rm 3}, 
 Guodong Guo\textsuperscript{ \rm 1,\rm 2 $\dag$
} 
\thanks{
Manuscript created November, 2021. {$*$} denotes equal contribution. (Corresponding author: Guodong Guo).

Sitong Wu, Tianyi Wu, Fangjian Lin, and Guodong Guo are with Institute of Deep Learning, Baidu Research, Beijing 100085, China, and also with National Engineering Laboratory for Deep Learning Technology and Application, Beijing 100085, China (email: wusitong@baidu.com; wutianyi01@baidu.com; linfangjian@baidu.com; guoguodong01@baidu.com). Fangjian Lin and Shengwei Tian are with School of Software, XinJiang University, urumqi, China. (email: tianshengwei@163.com).

}
}


\markboth{}
{S. Wu, \MakeLowercase{\textit{et al.}}: Fully Transformer Networks for Semantic Image Segmentation}

\maketitle

\begin{abstract}
Transformers have shown impressive performance in various natural language processing and computer vision tasks, due to the capability of modeling long-range dependencies. Recent progress has demonstrated that combining such Transformers with CNN-based semantic image segmentation models is very promising. However, it is not well studied yet on how well a pure Transformer based approach can achieve for image segmentation. In this work, we explore a novel framework for semantic image segmentation, which is encoder-decoder based Fully Transformer Networks (FTN). Specifically, we first propose a Pyramid Group Transformer (PGT) as the encoder for progressively learning hierarchical features, meanwhile reducing the computation complexity of the standard Visual Transformer (ViT). 
Then, we propose a Feature Pyramid Transformer (FPT) to fuse semantic-level and spatial-level information from multiple levels of the PGT encoder for semantic image segmentation. Surprisingly, this simple baseline can achieve better results on multiple challenging semantic segmentation and face parsing benchmarks, including PASCAL Context, ADE20K, COCO-Stuff, and CelebAMask-HQ. 
The source code will be released on \url{https://github.com/BR-IDL/PaddleViT}.
\end{abstract}


\section{Introduction}
\IEEEPARstart{S}{emantic} image segmentation aims to assign a semantic label to each pixel in an image, which is an indispensable component for many applications, such as autonomous vehicles \cite{Cityscapes}, 
human-computer interaction \cite{harders2003enhancing}, and medical diagnosis \cite{medical_diagnosis}. Since the era of deep learning, convolutional neural networks (CNNs) have been the cornerstone of tremendous image segmentation models.

\begin{figure*}[!t]
\centering
\includegraphics[width=0.75\linewidth]{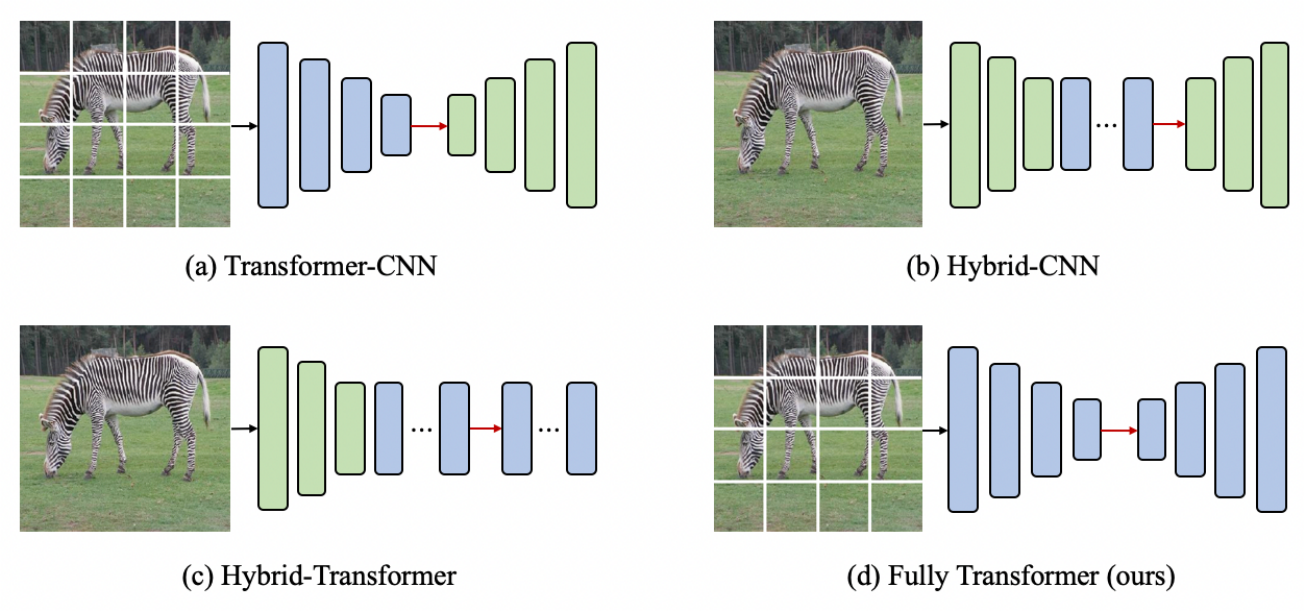}
\caption{Pipelines of the previous Transformer-based segmentation models. The blue bar represents Transformer blocks and the green bar indicates CNN layers. The encoder and decoder are separated by a red snip.}
\label{figure_1_1}
\end{figure*}

Inspired by successful applications of Transformer \cite{Transformer} on natural language processing tasks (machine translation \cite{machine_translation}, text classification \cite{text_classification}), and computer vision tasks (image classification \cite{VIT,DeiT,PVT,CvT, Swin, ViL, LocalViT}), Transformer has attracted more and more interests in the context of semantic image segmentation \cite{SETR, DPT, PVT, Swin, Trans2Seg}  recently. These methods combined the Transformer block into the encoder-decoder architecture, where the encoder reduces the feature maps and captures rich semantic representations, while the decoder gradually recovers the spatial information or fuses multi-scale features for generating accurate pixel-level predictions. These approaches can be divided into three categories: (1) \textbf{Transformer-CNN architecture} \cite{DPT, PVT,SETR, Swin}. It contains a Transformer encoder and CNN decoder. Images are first divided into a set of patches and projected to patch embeddings, which are then passed through a Transformer encoder and a traditional CNN decoder to recover the pixel-level prediction, as shown in Figure {\ref{figure_1_1}}(a). (2) \textbf{Hybrid-CNN architecture} \cite{SETR}. As shown in Figure {\ref{figure_1_1}}(b), its decoder is the traditional CNN, while the encoder is a combination of CNN (e.g., ResNet \cite{ResNet}) and Transformer layers, in which the output of CNN with a lower resolution is directly used as the patch embeddings for Transformer layers. (3) \textbf{Hybrid-Transformer architecture} \cite{Trans2Seg}. It further replaced the CNN decoder of Hybrid-CNN architecture with a Transformer decoder, as shown in Figure {\ref{figure_1_1}}(c).


Different from the architectures mentioned above, we propose Fully Transformer Networks (FTN) for semantic image segmentation, without relying on CNN. As shown in Figure {\ref{figure_1_1}}(d), both the encoder and decoder in the proposed FTN are composed of multiple Transformer layers. 
Firstly, we propose a new Vision Transformer encoder called Pyramid Group Transformer (PGT). Inspired by the hierarchical nature of the features in deep networks \cite{zeiler2014visualizing}), we propose to gradually increase the receptive fields rather than keep it global throughout the whole network (such as ViT \cite{VIT}). Specifically, PGT spatially divides the feature maps into multiple groups, and performs multi-head self-attention within each group. The receptive fields can be easily controlled by the group number, thus we progressively increase the number of groups, forming a pyramid pattern, to achieve gradually increasing receptive fields. 
Benefited by such design, our PGT is good at learning hierarchical features, including low-level (edge/color conjunctions), mid-level (complex invariances, similar textures), and high-level (entire objects).
Furthermore, PGT can also reduce the unaffordable computational and memory costs of the standard Visual Transformer (ViT), which is essential for pixel-level prediction tasks.
Secondly, we propose a Transformer-based decoder, named Feature Pyramid Transformer (FPT) to fuse semantic-level and spatial-level information from multiple levels of the PGT encoder. Benefited by the long-range modeling capacity of Transformer architecture, FPT can capture richer context information.
Based on the proposed encoder and decoder, we develop Fully Transformer Networks (FTN) for semantic image segmentation, and evaluated the effectiveness of our approach by conducting extensive ablation studies. 
Furthermore, we evaluate our FTN on three challenging
semantic segmentation benchmarks, including PASCAL-Context
\cite{PContext}, ADE20K \cite{ADE20K}, and COCO-Stuff \cite{caesar2018coco}, achieving the
state-of-the-art performance (mIoU) of 56.05\%, 51.36\%, and 45.89\%, respectively.

Our main contributions include:
\begin{itemize} 
\item We propose the Pyramid Group Transformer (PGT) to learn hierarchical representations by gradually increasing the receptive fields via a pyramid pattern. Meanwhile, it can reduce the computation and memory costs of the standard ViT significantly. 
\item We develop a Feature Pyramid Transformer (FPT) decoder to fuse semantic-level and spatial-level information from multiple levels of the PGT encoder. It is able to capture richer contextual information benefited by the long-range dependencies modeling capability of Transformer architecture.
\item Based on the proposed PGT and FPT, we create Fully Transformer Networks (FTN) for semantic image segmentation, which achieves superior performance than previous approaches on multiple challenging benchmarks, including PASCAL Context, ADE20K, COCO-Stuff, and CelebAMask-HQ.  
\end{itemize}

\section{RELATED WORKS}

\noindent 
\textbf{Semantic Image Segmentation.}
The early FCN \cite{FCN} used a fully convolutional network to generate dense predictions for images with an arbitrary size, which is regarded as a milestone in semantic segmentation. 
Since then, lots of works have been designing segmentation networks based on CNN architecture \cite{SegNet,DeepLabv2, PSPNet, CGNet, wu2019tree, 9334427, 8598722, MGSeg, GPSNet, 8760555, CGBNet, SAN}.
For more precise segmentation, some works \cite{DeconvNet, SegNet} were devoted to involving fine-grained details via various hierarchies of decoders. 
Considering that the effective receptive fields of CNN architecture are proved to be much smaller than its theoretical value \cite{effectiveRF}, some works focused on how to capture more contextual information for a better scene understanding\cite{DeepLabv2, DeepLabv3, DeepLabv3+, PSPNet, CGNet}. For example, DeepLab family \cite{DeepLabv2, DeepLabv3, DeepLabv3+} replaced partial pooling layers with atrous convolution to expand the receptive field without reducing the resolution. PSPNet \cite{PSPNet} proposed a general pyramid pooling module to fuse features under four different pyramid scales. CGNet \cite{CGNet} learns the joint features of both local features and the surrounding context in all stages of the network.
Recently, in order to aggregate more precise contextual information, \cite{OCRNet} and \cite{ContextEncoding} devote to capture class-dependent context information. \cite{ISNet,DependencyNet}, and \cite{CPNet} proposed to augment each pixel representation by aggregating intra-class and inter-class context respectively.
Besides, there are also some works committed to improving the segmentation on the boundary region \cite{InverseForm, BoundaryIoU, Segfix}.

\noindent 
\textbf{Self-attention in Semantic Image Segmentation.}
Inspired by the success of Transformer \cite{Transformer} for NLP tasks, 
DANet \cite{DANet} employed the self-attention \cite{Non-Local} on the CNN-based backbone, 
which adaptively integrated local features with their global dependencies. 
The traditional self-attention serves as a particularly useful technique for semantic segmentation while being criticized for its prohibitive computation burdens and memory usages. To address this issue, \cite{CCNet} and \cite{zhu2019asymmetric} introduced a criss-cross attention mechanism and pyramid sampling module into the self-attention, respectively, aiming to reduce the computation cost and memory usage while keeping the performance. AANet \cite{AANet} proposed an attention-augmented convolution architecture by concatenating convolutional feature maps with a set of feature maps produced via self-attention.
In contrast to these methods which augmented a CNN-based encoder with a self-attention module for semantic segmentation, we propose the novel Fully Transformer Networks, where both encoder and decoder are based on pure Transformer with our specialized design.

\noindent 
\textbf{Transformer in Vision Tasks.}
DETR \cite{DETR} combined a common CNN with Transformer for predicting objects, which formulates object detection as a dictionary lookup problem with learnable queries. Later, many researchers explored to adopt a pure Transformer architecture for computer vision tasks \cite{VIT, DeiT, PVT, Swin, CvT, LocalViT, ViL, MViT}. 
Vision Transformer (ViT) \cite{VIT} is the first work to introduce a pure Transformer into image classification by treating an image as a sequence of patches. Recently, Transformer has attracted more and more attention in semantic segmentation \cite{SETR, Trans2Seg, DPT}. SETR \cite{SETR} attempted to introduce Transformer for semantic segmentation by using ViT as the encoder and employing CNN-based decoders for generating segmentation results. Trans2Seg \cite{Trans2Seg} stacked a convolution-free encoder-decoder network on top of a common CNN encoder for transparent object segmentation. DPT \cite{DPT} assembled tokens from multiple stages of ViT and progressively combined them into full-resolution predictions using a convolutional decoder. Different from these approaches that used Transformer as an encoder or decoder, we proposed the encoder-decoder based Fully Transformer Networks for semantic segmentation.

\begin{figure*}[!h]
	\centering
		\includegraphics[width=0.85\linewidth]{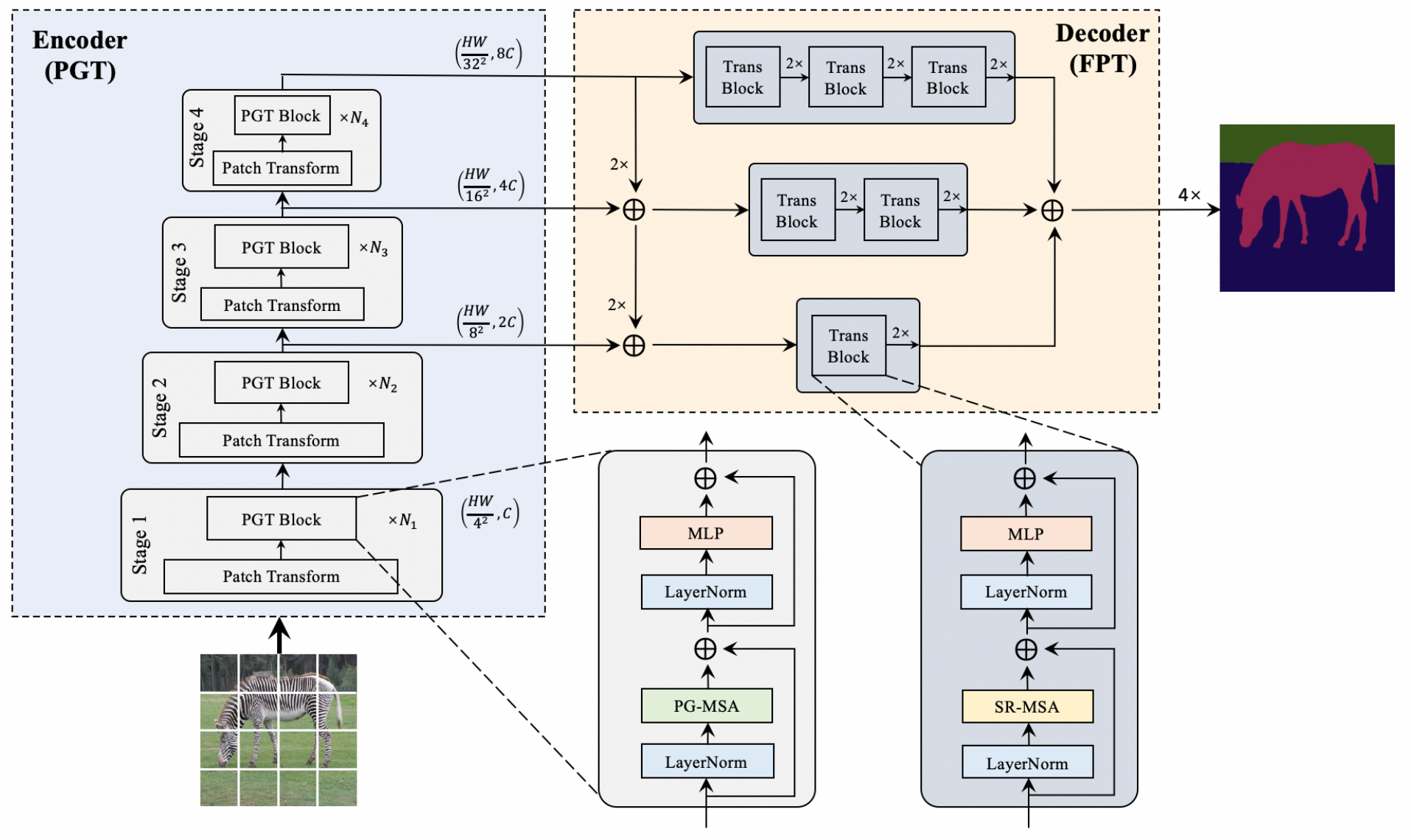}
		\vspace{-0.2cm}
		\caption{ \label{Framework}The overall architecture of our Fully Transformer Networks (FTN). Both the encoder and decoder of FTN are based on Transformer architectures.}
\end{figure*}

\section{METHOD}
In this section, we first describe our framework, Fully Transformer Networks (FTN). Then, we present the encoder architecture of FTN, \emph{i.e.}, Pyramid Group Transformer (PGT), which aims to extract hierarchical representations. Finally, we present the decoder architecture of FTN, \emph{i.e.}, Feature Pyramid Transformer (FPT), which fuses the multi-scale features from the encoder and generates the fine pixel-level predictions.

\subsection{Framework of FTN}
The overall framework of our FTN is shown in Figure \ref{Framework}, which consists of Pyramid Group Transformer (PGT) encoder and Feature Pyramid Transformer (FPT) decoder. 
PGT aims to extract hierarchical representations. 
It is configured with four stages for learning hierarchical features, 
similar to some classic backbones \cite{ResNet,xie2017aggregated}. Each stage of PGT has a similar structure, which contains a patch transform layer and multiple PGT blocks. The patch transform layer is employed to reduce the number of tokens.
In particular, given an input image $x\in\mathbb{R}^{H\times{W}\times{3}}$, it is first transformed into $\frac{HW}{4^{2}}$ patches with dimension $C$ by the patch transform layer in stage 1, then the output is fed into $N_{1}$ PGT blocks, where $N_1$ indicates the number of PGT blocks in stage 1. The output of the last block in stage 1 is $F_1\in\mathbb{R}^{\frac{HW}{4^{2}}\times{C}}$. For the last $3$ stages, the patch transform layer merges each 2 × 2 non-overlapping neighboring patches to reduce the resolution by 1/2 and expand the dimension twice. The output feature in stage $i$ is $F_i\in\mathbb{R}^{\frac{HW}{2^{2i+2}}\times{2^{i-1}C}}$. 
After getting multi-scale features, we employ FPT decoder to fuse semantic-level and spatial-level information from multiple levels. Finally, the output of FPT is fed into a linear layer followed by a simple bilinear upsampling to generate the probability map for each pixel. The pixel-level segmentation result is obtained by argmax operation on the probability map.


\begin{figure}[!h]
\centering
\subfloat[ViT] {\label{SA_1}\includegraphics[width=0.2\linewidth]{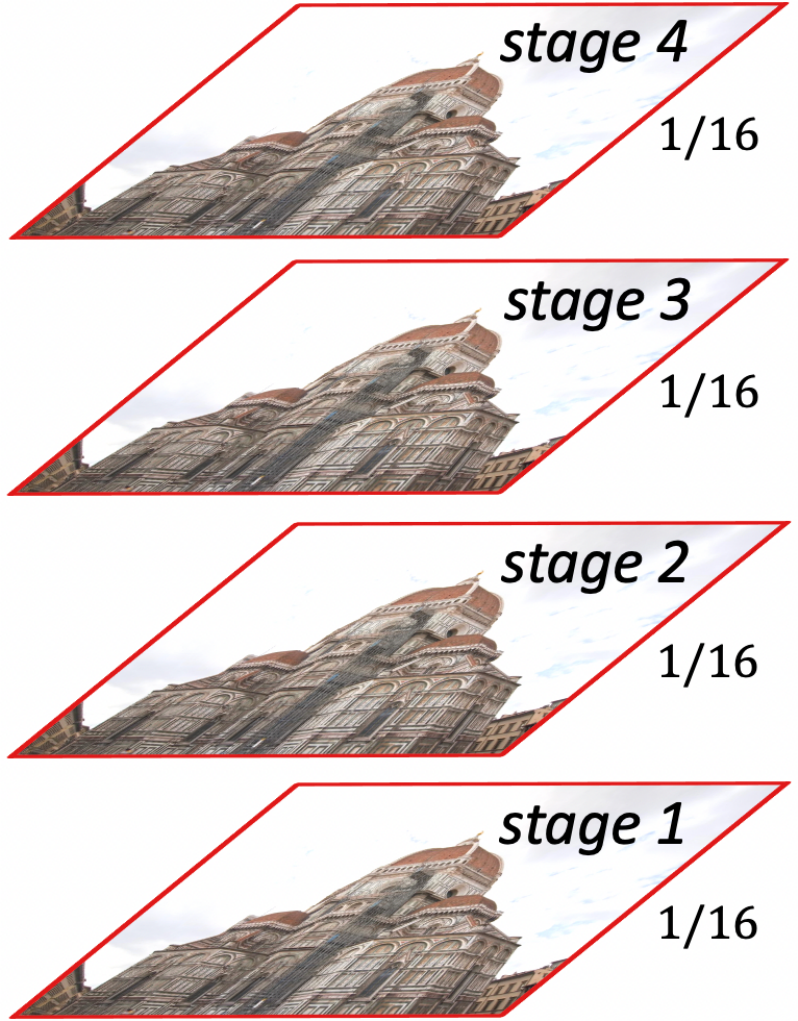}} 
\hfil
\subfloat[PVT] {\label{SA_2}\includegraphics[width=0.32\linewidth]{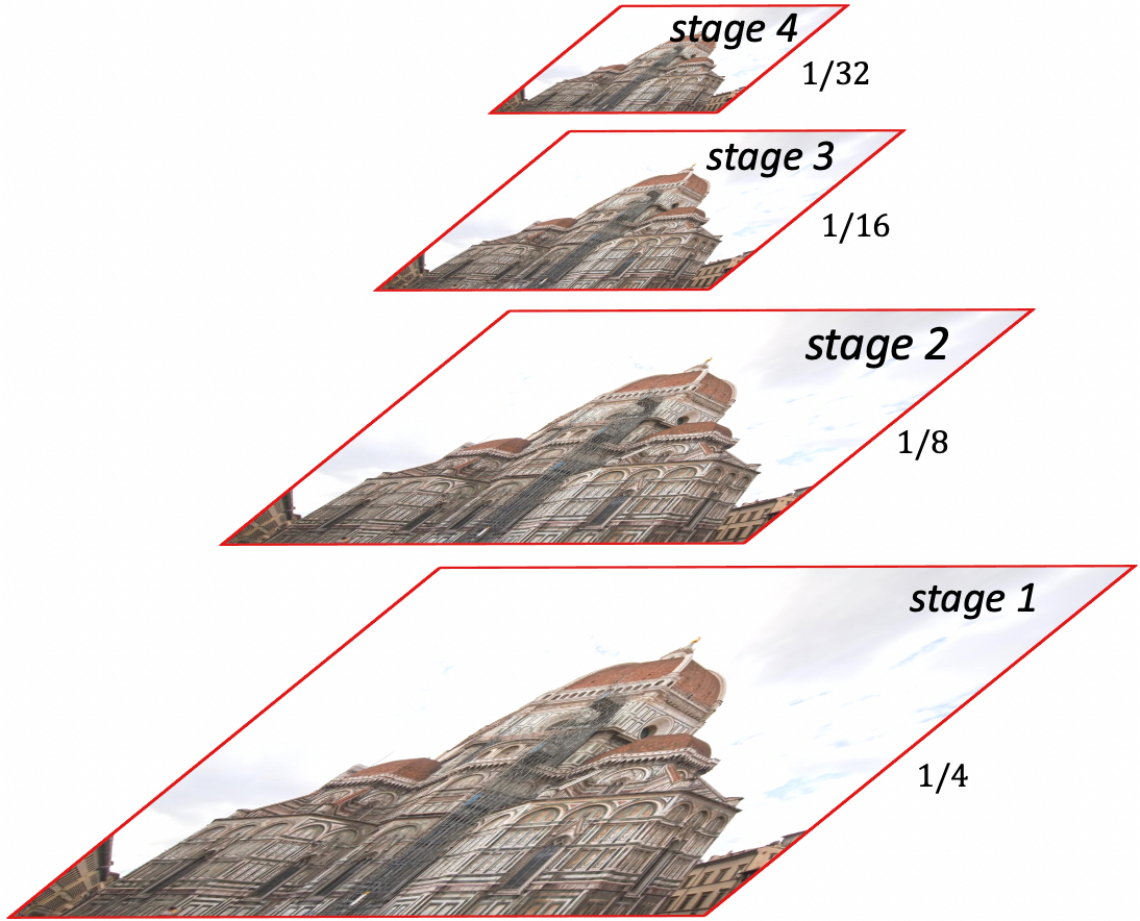}} 
\hfil
\subfloat[PGT (ours)] {\label{SA_3}\includegraphics[width=0.32\linewidth]{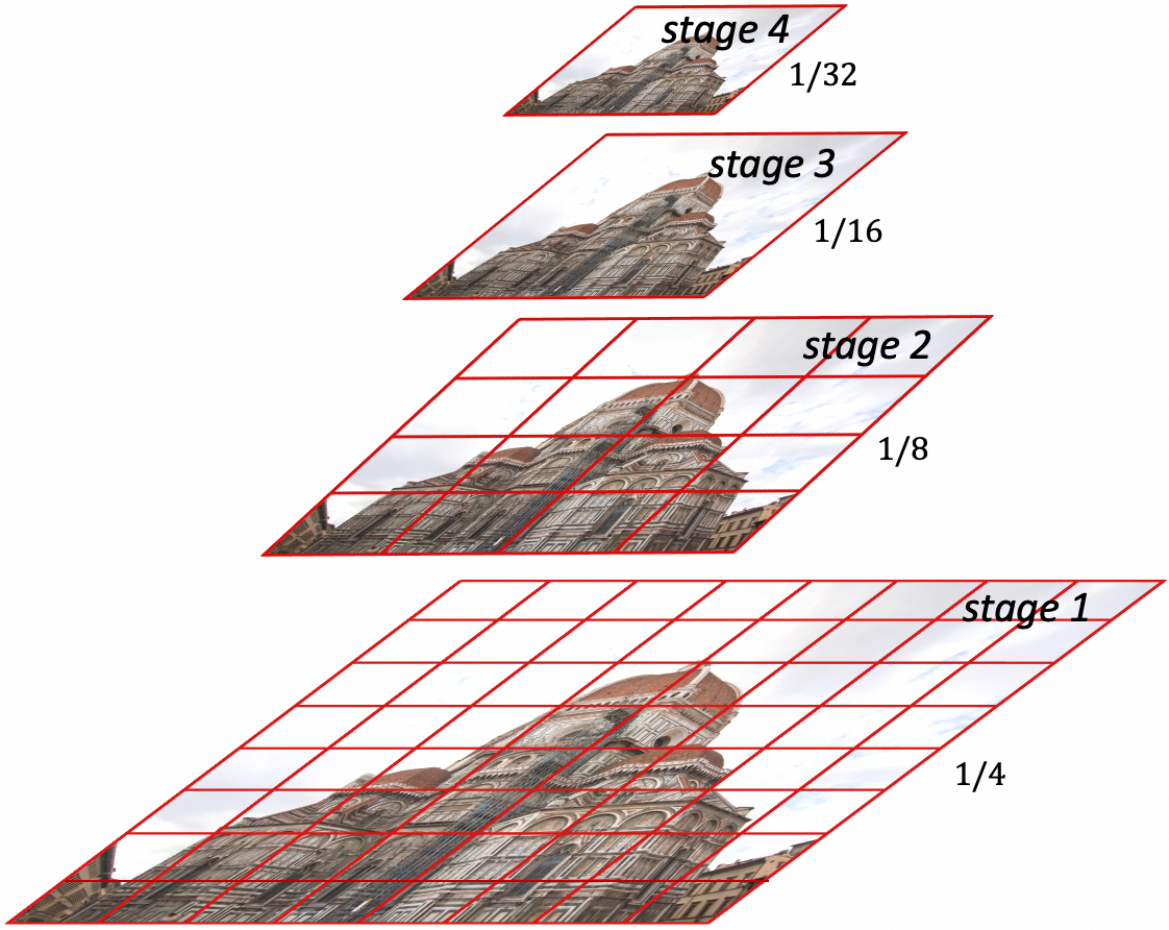}} 
\caption{ \label{SA} Comparisons of different self-attention mechanisms in Transformers. (a) ViT performed the global self-attention within the single-scale feature maps throughout the whole network. (b) PVT introduced the pyramid structure to generate multi-scale feature maps and applied the global self-attention to features as well. (c) In PGT, the feature map is first divided into groups in the spatial domain and then the global self-attention is only performed within each group, i.e., a red grid in the figure. The receptive fields (group size) of attention are progressively increased in stages as a pyramid pattern.}
\end{figure}

\begin{table*}[tp]
\caption{Detailed configuration of our PGT variants.}
\label{Variants}
\newcommand{\tabincell}[2]{\begin{tabular}{@{}#1@{}}#2\end{tabular}}
\renewcommand{\arraystretch}{1.0}
\centering
\resizebox{0.75\textwidth}{!}{
\begin{tabular}{ c | c | c | c | c | c | c }
\toprule[1pt]
\multirow{2}{*}{Stage}    & \multirow{2}{*}{\tabincell{c}{Output \\ Stride}}    & \multirow{2}{*}{Layer}    & \multirow{2}{*}{PGT-T}     & \multirow{2}{*}{PGT-S}      & \multirow{2}{*}{PGT-B}    & \multirow{2}{*}{PGT-L} \\
&  &    &   &       &   & \\
\midrule[1pt]
\multirow{6.5}{*}{1}    & \multirow{6.5}{*}{4}    & \multirow{2}{*}{\tabincell{c}{Patch\\Transform}}     & \multirow{2}{*}{\tabincell{c}{$P_1=4$\\$C_1=64$}}     & \multirow{2}{*}{\tabincell{c}{$P_1=4$\\$C_1=96$}}   & \multirow{2}{*}{\tabincell{c}{$P_1=4$\\$C_1=96$}}   & \multirow{2}{*}{\tabincell{c}{$P_1=4$\\$C_1=128$}}   \\ 
&  &    &   &       &   & \\
\cmidrule{3-7}
& &\multirow{4}{*}{\tabincell{c}{PGT \\ Block}}   & \multirow{4}{*}{$\begin{bmatrix}  G_1=64 \\ H_1=2 \\ E_1=4 \end{bmatrix} \times 2$}    &  \multirow{4}{*}{$\begin{bmatrix}  G_1=64 \\ H_1=3 \\ E_1=4 \end{bmatrix} \times 2$}   &  \multirow{4}{*}{$\begin{bmatrix}  G_1=64 \\ H_1=3 \\ E_1=4 \end{bmatrix} \times 2$}   & \multirow{4}{*}{$\begin{bmatrix}  G_1=64 \\ H_1=4 \\ E_1=4 \end{bmatrix} \times 2$}   \\ 
& &    &   &       &     &  \\
&  &    &   &       &    & \\
&  &    &   &       &    & \\
\midrule[0.5pt]
\multirow{6.5}{*}{2}   & \multirow{6.5}{*}{8}        & \multirow{2}{*}{\tabincell{c}{Patch\\Transform}}     & \multirow{2}{*}{\tabincell{c}{$P_2=2$\\$C_2=128$}}     & \multirow{2}{*}{\tabincell{c}{$P_2=2$\\$C_2=192$}}  & \multirow{2}{*}{\tabincell{c}{$P_2=2$\\$C_2=192$}}     & \multirow{2}{*}{\tabincell{c}{$P_2=2$\\$C_2=256$}}   \\ %
&  &    &   &       &    & \\
\cmidrule{3-7}
& & \multirow{4}{*}{\tabincell{c}{PGT \\ Block}}   & \multirow{4}{*}{$\begin{bmatrix}  G_2=16 \\ H_2=4 \\ E_2=4 \end{bmatrix} \times 2$}    & \multirow{4}{*}{$\begin{bmatrix}  G_2=16 \\ H_2=6 \\ E_2=4 \end{bmatrix} \times 2$}    & \multirow{4}{*}{$\begin{bmatrix}  G_2=16 \\ H_2=6 \\ E_2=4 \end{bmatrix} \times 2$}   & \multirow{4}{*}{$\begin{bmatrix}  G_2=16 \\ H_2=8 \\ E_2=4 \end{bmatrix} \times 2$}   \\ 
&  &    &   &       &   &  \\
&  &    &   &       &   &  \\
&  &    &   &       &    & \\
\midrule[0.5pt]          
\multirow{6.5}{*}{3}   & \multirow{6.5}{*}{16}        & \multirow{2}{*}{\tabincell{c}{Patch\\Transform}}   & \multirow{2}{*}{\tabincell{c}{$P_3=2$\\ $C_3=256$}}     & \multirow{2}{*}{\tabincell{c}{$P_3=2$\\$C_3=384$}}   & \multirow{2}{*}{\tabincell{c}{$P_3=2$\\$C_3=384$}}   & \multirow{2}{*}{\tabincell{c}{$P_3=2$\\$C_3=512$}}   \\ %
&  &    &   &       &    & \\
\cmidrule{3-7}
& & \multirow{4}{*}{\tabincell{c}{PGT \\ Block}}   & \multirow{4}{*}{$\begin{bmatrix}  G_3=1 \\ H_3=8 \\ E_3=4 \end{bmatrix} \times 6$}    & \multirow{4}{*}{$\begin{bmatrix}  G_3=1 \\ H_3=12 \\ E_3=4 \end{bmatrix} \times 6$}  & \multirow{4}{*}{$\begin{bmatrix}  G_3=1 \\ H_3=12 \\ E_3=4 \end{bmatrix} \times 18$}  & \multirow{4}{*}{$\begin{bmatrix}  G_3=1 \\ H_3=16 \\ E_3=4 \end{bmatrix} \times 18$}   \\ 
&  &    &   &       &   &  \\
&  &    &   &       &   &  \\
&  &    &   &       &   &  \\
\midrule[0.5pt]    
\multirow{6.5}{*}{4}   & \multirow{6.5}{*}{32}        & \multirow{2}{*}{\tabincell{c}{Patch\\Transform}}         & \multirow{2}{*}{\tabincell{c}{$P_4=2$\\$C_4=512$}}     & \multirow{2}{*}{\tabincell{c}{$P_4=2$\\$C_4=768$}}    & \multirow{2}{*}{\tabincell{c}{$P_4=2$\\$C_4=768$}}   & \multirow{2}{*}{\tabincell{c}{$P_4=2$\\$C_4=1024$}}   \\ %
&  &    &   &       &   &   \\
\cmidrule{3-7}
& & \multirow{4}{*}{\tabincell{c}{PGT \\ Block}}   & \multirow{4}{*}{$\begin{bmatrix}  G_4=1 \\ H_4=16 \\ E_4=4 \end{bmatrix} \times 2$}    & \multirow{4}{*}{$\begin{bmatrix}  G_4=1 \\ H_4=24 \\ E_4=4 \end{bmatrix} \times 2$}    & \multirow{4}{*}{$\begin{bmatrix}  G_4=1 \\ H_4=24 \\ E_4=4 \end{bmatrix} \times 2$}    & \multirow{4}{*}{$\begin{bmatrix}  G_4=1 \\ H_4=32 \\ E_4=4 \end{bmatrix} \times 2$}   \\ 
&  &    &   &       &  &   \\
&  &    &   &       &   &  \\
&  &    &   &       &   &  \\
\bottomrule[1pt]    
\end{tabular}}            
\end{table*}

\subsection{Pyramid Group Transformer}
\label{PGA}

As shown in Figure \ref{Framework}, PGT has four hierarchical stages that generate features with multiple scales. At the beginning of each stage, the features are first shrunk spatially and enlarged in channel dimension by the patch transform layer, and then be fed into the subsequent PGT blocks to learn discriminative representations.
PGT progressively increases the receptive field of self-attention when the stage increases, so as to learn low-level spatial details at shallow layers and high-level semantic features at deep layers. Such a mechanism is superior to the previous Vision Transformer (ViT) \cite{VIT} where the self-attention with a global receptive field is performed on the whole input features at all layers. Besides, our approach can reduce the computational and memory cost of the whole transformer block, which is desired especially for dense prediction tasks.

Specifically, each feature map is firstly divided into non-overlapping grids and each grid is regarded as a group. Then, self-attention is performed among patches within each group. Thus, patches in one group are unacceptable for patches in other groups, which is equivalent to local receptive fields. Explicitly, the size of each receptive field can be conveniently controlled by the group numbers. As shown in Figure \ref{SA}, different from the global receptive fields in all stages of ViT \cite{VIT} and PVT \cite{PVT}, the receptive fields at different stages of our PGT present a pyramid pattern, and we set a consistent receptive field inside each stage. It is notable that the global receptive field is better to be applied in both stage 3 and stage 4 for the reason that features with stride 16 and 32, usually contain rich semantics. For the $l$-th PGT block, its computation can be formulated as follows:
\begin{equation} 
    \begin{aligned}
    &\hat{Z}^l = Z^{l-1} + {\rm PG\verb|-|MSA}({\rm LN}(Z^{l-1})), \\
    &Z^l = \hat{Z}^l + {\rm MLP}({\rm LN}(\hat{Z}^l)),
    \end{aligned}
\end{equation}
where $Z^{l-1}$ is the output of the $(l-1)$-th PGT block. LN and MLP refer to layer normalization \cite{ba2016layer} and multi-layer perceptron, respectively. 
Furthermore, the core of PGT blocks is the Pyramid Group Multi-Self Attention (PG-MSA), which can be formulated as follow:
\begin{equation} 
    \begin{aligned}
    & {\rm PG\verb|-|MSA}(Z)= {\rm Concat}( h_0, h_1, h_{H-1}), \\
    & h_i = {\rm Reshape} (head_{i}^{0},head_{i}^{1},..., head_{i}^{G-1}),  \\
    & head_{i}^{j}= {\rm Attention}(Q_i[j],K_i[j],V_i[j]),
    \end{aligned}
\end{equation}
where $i\in\{0,1,...,H-1\}$ is the head index; $j\in \{0,1,...,G-1\}$ is the group index; 
${\rm Attention}(\cdot)$ is the self-attention operation \cite{Transformer}.
$Q_i=ZW_i^q$, $K_i=ZW_i^k$, and $V_i=ZW_i^v$ indicate the query, key, value embedding of the $i$-th head, respectively.

\noindent 
\textbf{PGT Variants.}
We designed 4 variants with different sizes, named PGT-T, PGT-S, PGT-B, and PGT-L respectively. For a fair comparison, the sizes of these variants are similar to the previous works \cite{PVT, Swin, DeiT}. 
The detailed configuration of our PGT variants are shown in Table \ref{Variants}, where the definitions of hyper-parameters are listed below:

\begin{itemize}
\setlength{\itemsep}{0pt}
\setlength{\parsep}{0pt}
\setlength{\parskip}{0pt}
\item ${P_i}^2$: the token number reduction factor of the patch transform layer in stage $i$;
\item $C_i$: the dimension of tokens in stage $i$;
\item $N_i$: the number of PGT blocks in stage $i$;
\item $G_i$: the group number of PG-MSA in stage $i$;
\item $H_i$: the head number of PG-MSA in stage $i$;
\item $E_i$: the dimension expansion ratio of MLP in stage $i$;
\end{itemize}

According to the differences among these variants, the representation capacity is only enlarged by increasing the depth of PGT blocks and the dimension of patch tokens (i.e. deeper and wider). For example, PGT-S is obtained by increasing the dimensions of PGT-T, and PGT-B is a deeper version of PGT-S. PGT-L is the deepest and widest among these all. For all variants, the expansion ratio of MLP is set to 4, and the dimension of each head is 32. The pyramid structure of the receptive fields in attention is also consistent for different variants, with group numbers 64, 16, 1, and 1, respectively, in four stages.

\noindent 
\textbf{Compared with Swin Transformer.}
For Swin Transformer \cite{Swin}, the window size is set as a fixed number for input image with arbitrary size. While, in our PGT, the group number is constant for each stage and gradually decreases from shallow stage to deep stage. The group size depends on the size of the input image. Therefore, as the image size increases, the receptive field of PGT enlarges faster than that of Swin.

\noindent 
\textbf{Compared with PVT.}
PVT \cite{PVT} proposed Spatial Reduction Attention (SRA) module to reduce the computational and memory cost of global attention, which makes it affordable for high-resolution features. Nevertheless, the receptive fields in all stages of PVT are still global. Different from that, we propose to gradually increase the receptive fields rather than global throughout the network. Thus, our PGT does better in learning low-level features with detailed information, and high-level features with rich semantics.


\subsection{Feature Pyramid Transformer}
To generate finer semantic image segmentation, we propose a Feature Pyramid Transformer (FPT) to aggregate the information from multiple levels of the PGT encoder, as shown in Figure \ref{Framework}. 
Inspired by the spirit of convolution-based FPN \cite{FPN}, FPT is expected to fuse the semantic-level representations and spatial-level information into a high-resolution and high-level semantic output. 

Specifically, given the output hierarchical features $\{F_i\}, i \in \{1,2,...,L\}$ of the encoder, we first use a light-weight fusion module, containing lateral connections and a top-down pathway, to introduce strong semantics into all levels, 
\begin{equation} 
\begin{aligned}
\widehat{F}_i = 
\begin{cases}
\phi(F_i) + {\rm Upsample} \Big( \widehat{F}_{i+1} \Big), & 1 \leq i < L \\
\phi(F_i), & i = L  
\end{cases}
\end{aligned}
\end{equation}
where $\phi$ is used to align the channel dimension, implemented by a linear projection layer. ${\rm Upsample}$ denotes bilinear interpolation. Note that $\widehat{F}_i$ has the same size of $F_i$. Then, each level feature is passed through the spatial reduction Transformer blocks \cite{PVT} and progressively upsampled until it reaches $1/4$ scale, 
\begin{equation} 
\begin{aligned}
\widetilde{F}_i = 
\begin{cases}
\Psi(\widehat{F}_i), & i = 1  \\
\Big [ {\rm Upsample} \Big( \Psi(\widehat{F}_i) \Big) \Big ] _{\times(i-1)}, & 1 < i \leq L
\end{cases}
\end{aligned}
\end{equation}
where $\Psi$ represents Transformer blocks and ${\rm Upsample}$ is implemented by bilinear interpolation. $[ \ f \ ]_{\times N}$ denotes executing function $f$ by $N$ times. Finally, these multi-level and high-resolution representations are fused by simple element-wise summation, and further passed through a per-pixel classification head to predict pixel-level prediction and then upsampled to the original size of input image via bilinear interpolation.
\begin{equation} 
\begin{aligned}
& Y = {\rm Upsample} \bigg( \psi \Big( \sum_{i=1}^{L} \widetilde{F}_i \Big) \bigg), 
\end{aligned}
\end{equation}
where $\psi$ denotes per-pixel classification head, which is implemented by a linear projection layer. Benefited by the top-down connections and multi-level aggregation, our model is able to enhance the capability of fusing multi-level semantic features with different resolutions, which is essential for dense prediction tasks.


\label{experiments}
\section{Experiments}

In this section, we first compare our PGT with the state-of-the-art methods on the image classification task. To further demonstrate the effectiveness of our FTN, we conduct experiments on three widely-used segmentation benchmarks, PASCAL Context \cite{PContext}, ADE20K \cite{ADE20K}, and COCO-Stuff \cite{caesar2018coco}. Detailed ablation studies are also provided to understand the importance and impact of individual components and settings in our method.


\begin{table}[h]
\caption{\label{CLS_EXP} Comparisons of different backbones for the ImageNet-1K classification. Our PGT consistently outperforms other models with similar numbers of parameters and computational budgets.}
\centering
\scalebox{0.9}{
\begin{tabular}{c|ccc|c}
\toprule[1pt]
\textbf{Method}   & \textbf{Input size} & \textbf{Params}    & \textbf{FLOPs}   & \textbf{Top-1 Acc. (\%)} \\
\midrule[1pt] 
ResNet-18 \cite{ResNet}         & 224     & 12M     & 1.8G     & 68.5     \\
PVT-T \cite{PVT}             & 224     & 13M     & 1.9G     & 75.1     \\
\textbf{PGT-T (ours)}                     & 224     & 13M     & 2.1G      & \textbf{79.9}    \\ 
\midrule[0.5pt] 
ResNet-50 \cite{ResNet}	        & 224     & 26M     & 4.1G     & 78.5     \\
DeiT-S \cite{DeiT}          & 224     & 22M     & 4.6G     & 79.8     \\
PVT-S \cite{PVT}            & 224     & 25M     & 3.8G     & 79.8     \\
RegNetY-4G \cite{RegNet}        & 224     & 21M     & 4.0G     & 80.0     \\
T2T-ViT$_t$-14 \cite{T2TVIT}    & 224     & 22M     & 6.1G     & 80.7     \\
TNT-S \cite{TNT}                & 224     & 24M     & 5.2G     & 81.3     \\ 
Swin-T \cite{Swin}              & 224     & 29M     & 4.5G     & 81.3     \\
CvT-13 \cite{CvT}               & 224     & 20M     & 4.5G     & 81.6     \\
\textbf{PGT-S (ours)}                     & 224     & 28M     & 4.6G     & \textbf{82.0}    \\ 
\midrule[0.5pt]
ResNet-101 \cite{ResNet}	    & 224     & 45M     & 7.9G     & 79.8     \\
PVT-M \cite{PVT}           & 224     & 44M     & 6.7G     & 81.2     \\
T2T-ViT$_t$-19 \cite{T2TVIT}    & 224     & 39M     & 9.8G     & 81.4     \\ 
RegNetY-8G \cite{RegNet}        & 224     & 39M     & 8.0G     & 81.7     \\
CvT-21 \cite{CvT}               & 224     & 32M     & 7.1G     & 82.5     \\
Swin-S \cite{Swin}              & 224     & 50M     & 8.7G     & 83.0     \\
\textbf{PGT-B (ours)}                     & 224     & 50M     & 9.1G     & \textbf{83.4}    \\ 
\midrule[0.5pt]
PVT-L \cite{PVT}            & 224     & 61M     & 9.8G     & 81.7      \\
DeiT-B \cite{DeiT}           & 224     & 86M     & 17.5G    & 81.8      \\
T2T-ViT$_t$-24 \cite{T2TVIT}    & 224     & 64M     & 15.0G    & 82.2     \\ 
TNT-B \cite{TNT}                & 224     & 66M     & 14.1G    & 82.8     \\ 
RegNetY-16G \cite{RegNet}       & 224     & 84M     & 16.0G    & 82.9      \\
Swin-B \cite{Swin}              & 224     & 88M     & 15.4G    & 83.3      \\
\textbf{PGT-L (ours)}                     & 224     & 88M     & 15.9G    & \textbf{83.6}     \\
\midrule[0.5pt]
ViT-B/16 \cite{VIT}             & 384     & 86M     & 55.4G    & 77.9      \\
ViT-L/16 \cite{VIT}             & 384     & 307M    & 190.7G   & 76.5      \\
\bottomrule[1pt]
\end{tabular}}
\end{table}

\subsection{Image Classification}
In order to train our FTN on semantic image segmentation benchmarks, we first pretrained our PGT on the ImageNet dataset \cite{ImageNet}  for obtaining pretrained weights. ImageNet dataset contains 1.3 million images with 1,000 classes. It is referred to as ImageNet-1K in this work.

\label{CLS_setup}
\noindent 
\textbf{Experiment Settings.}
All the variants are trained for 300 epochs on 8 V100 GPUs from scratch, with a total batch size of 512. We use AdamW \cite{AdamW} as the optimizer with an initial learning rate of 0.0005, cosine decay scheduler, weight decay 0.05, and 5-epoch linear warmup. For data augmentation, we follow most of the widely used settings: random horizontal flipping \cite{H_flip}, color jitter, Mixup \cite{Mixup} and AutoAugment \cite{AutoAugment} and randomly cropping to 224 × 224. We also use some regularization strategies in \cite{DeiT}, such as Label-Smoothing \cite{Label_Smoothing} and stochastic depth \cite{Drop_Path}. During the evaluation, the input image is resized and center-cropped to the size of 224 × 224. The top-1 accuracy is reported for comparisons.

\noindent 
\textbf{Results on ImageNet-1K.}
Table \ref{CLS_EXP} shows the comparison with the state-of-the-art CNN and Transformer backbones. Compare to the dominant CNNs, our PGT consistently performs better than the state-of-the-art RegNet \cite{RegNet} and conventional ResNet \cite{ResNet} with similar complexity by a large margin. For example, our PGT-S is +2\% higher than RegNetY-4G and +3.5\% higher than ResNet-50. The advantage of PGT is especially obvious on tiny models, with PGT-T +4.8\% higher than PVT-T. For small variants, our PGT-S surpasses the nearest CvT-13 and Swin-T by +0.4\% and +0.7\%, respectively. Compared with larger models, our PGT-B and PGT-L exceed PVT significantly by 1$\sim$2\%, and outperform the state-of-the-art Swin Transformer with similar numbers of parameters and computation budgets, \emph{i.e.,} 83.6 \emph{vs.} 83.3 and 83.4 \emph{vs.} 83.0.


\begin{table*}[tp]
  \caption{\label{SEG_EXP} Comparisons with the state-of-the-art segmentation models on PASCAL Context, ADE20K and COCO-Stuff datasets. All the mIoU is obtained by multi-scale inference. ``$\ddagger$'' indicates pretraining on larger ImageNet-21k dataset under the input size of $384 \times 384$. ``$-$'' means no public results available.}
  \centering
  \resizebox{0.76\textwidth}{!}{
  \begin{tabular}{c|cc|ccc}
    \toprule[1pt]
    \multirow{2}{*}{\textbf{Method}}    & \multicolumn{2}{c|}{\textbf{Backbone}} & \multicolumn{3}{c}{\textbf{mIoU}}             \\ 
    & \textbf{Name}  & \textbf{Params}  & \textbf{PASCAL Context}     & \textbf{ADE20K}    & \textbf{COCO-Stuff} \\ 
    \midrule[1pt] 
    FCN \cite{FCN}                & ResNet-101 \cite{ResNet}  & 45M          & 45.63  & 41.40  & -       \\
    CGBNet\cite{CGBNet} & ResNet-101 \cite{ResNet}         & 45M  & 53.40 & 44.90 & 37.70 \\
    PSPNet \cite{PSPNet}          & ResNet-101 \cite{ResNet}         & 45M     & 47.78  & 45.35  & -       \\
    DeepLabV3+ \cite{DeepLabv3+}  & ResNet-101 \cite{ResNet}         & 45M      & 48.47  & 46.35  & -       \\
    CFNet\cite{wu2021consensus}  & ResNet-101 \cite{ResNet}         & 45M  & 52.4  & - & 36.6  \\
    DANet \cite{DANet}            & ResNet-101 \cite{ResNet}         & 45M        & 52.60  & 45.02  & 39.70   \\
    OCRNet \cite{OCRNet}          & ResNet-101 \cite{ResNet}          & 45M       & 54.80  & 45.28  & 39.50   \\
    Efficient-FCN\cite{liu2020efficientfcn}  & ResNet-101 \cite{ResNet}  & 45M    & 53.30  & 45.28  & -       \\
    GINet \cite{GINet}            & ResNet-101 \cite{ResNet}          & 45M       & 54.90  & 45.54  & 40.60   \\
    RecoNet \cite{chen2020tensor}  & ResNet-101 \cite{ResNet}         & 45M        &54.80   & 45.54   & 41.50  \\
    GPSNet\cite{GPSNet} & ResNet-101 \cite{ResNet}         & 45M   & - & 45.76 & - \\
    FPT\cite{FPT}  & ResNet-101 \cite{ResNet}         & 45M   & -  & 45.90 & -  \\
    CPNet\cite{CPNet} & ResNet-101 \cite{ResNet}  & 45M   & 53.9 & 46.27 & - \\
    ISNet\cite{ISNet} & ResNet-101 \cite{ResNet}         & 45M   & - & 47.55 & 42.08 \\
    \midrule[0.5pt] 
    UperNet \cite{upernet}        & Swin-T \cite{Swin}                & 28M                & 48.99  & 45.81  & 39.13    \\
    UperNet \cite{upernet}        & Swin-S \cite{Swin}                & 50M                  & 51.66  & 49.47  & 41.58    \\
    UperNet \cite{upernet}        & Swin-B \cite{Swin}                & 88M                 & 52.57  & 49.72  & 42.20    \\
    SETR-MLA \cite{SETR}          & ViT-L/16 $^{\ddagger}$ \cite{VIT}   & 307M   & 55.83  & 50.28  & -    \\
    \midrule[0.5pt] 
    FTN (ours)                    & PGT-T (ours)                       & 13M       & 51.15   & 47.12  & 41.57 \\
    FTN (ours)                    & PGT-S (ours)                       & 28M      & 53.09  & 48.68  & 43.63    \\
    FTN (ours)                    & PGT-B (ours)                       & 50M         & 54.93  & 50.88 & 44.82    \\
    FTN (ours)                    & PGT-L (ours)                       & 88M         & 56.05  & 51.36  & 45.89    \\
    \bottomrule[1pt] 
    \end{tabular}
    }
\end{table*}


\subsection{Semantic Image Segmentation}

We evaluate the segmentation performance of our FTN on PASCAL Context \cite{PContext}, ADE20K \cite{ADE20K}, and COCO-Stuff \cite{caesar2018coco} datasets.

\noindent 
\textbf{Datasets.} 
PASCAL Context \cite{PContext} is an extension of the PASCAL VOC 2010 detection challenge. Due to the sparsity of some categories, a subset of 60 classes is more commonly used. For fair comparisons, we also use this 60 classes (59 classes and background) subset, which contains 4998 and 5105 images for training and validation, respectively. ADE20K \cite{ADE20K} is a more challenging scene parsing dataset, which is split into 20210, 2000, and 3352 images for training, validation and testing, respectively, with 150 fine-grained object categories. COCO-Stuff-10K \cite{caesar2018coco} has 9000 training images and 1000 testing images with 182 categories, which is referred as COCO-Stuff in this paper.

\label{SEG_setup}
\noindent 
\textbf{Experimental Details.} We optimize our models using AdamW \cite{AdamW} with a batch size of 16. The total iterations are set to 80k, 160k and 100k for PASCAL Context, ADE20K, and COCO-Stuff, respectively. The initial learning rate is set to 6e-5 with a polynomial decay scheduler and weight decay 0.01. During training, data augmentation in all the experiments consists of three steps: (i) random horizontal flipping, (ii) random scale with factors between 0.5 and 2, (iii) random cropping 480 for PASCAL Context and 512 for ADE20K and COCO-Stuff. During inference, the multi-scale (MS) (factors vary from 0.5 to 1.75 with 0.25 as interval) mIoU is reported. For a fair comparison, we simply apply the cross-entropy loss and synchronized BN. Similar to Swin \cite{Swin}, we add auxiliary loss to the output of stage 3 of PGT with the weight of 0.4.

\noindent 
\textbf{Results on PASCAL Context.} 
Table \ref{SEG_EXP} shows the results on PASCAL Context dataset. Our FTN models with PGT variants as the encoder are significantly superior to the corresponding UperNet(Swin) models with similar numbers of parameters and computation cost. For example, our FTN(PGT-S) is +4.1\% higher than UperNet(Swin-T) (53.09 \emph{vs.}48.99). Our FTN(PGT-L) achieves 56.05\% mIoU, outperforming the UperNet(Swin-B) with 52.57\% mIoU by a large margin. 
Surprisingly, our FTN(PGT-L) which is pretrained on ImageNet-1K even surpasses the SETR-MLA(ViT-L/16) by +0.22\%, whose encoder pretrained on a larger ImageNet-21K dataset. 


\begin{figure}[!h]
	\centering
	\includegraphics[width=0.98\linewidth]{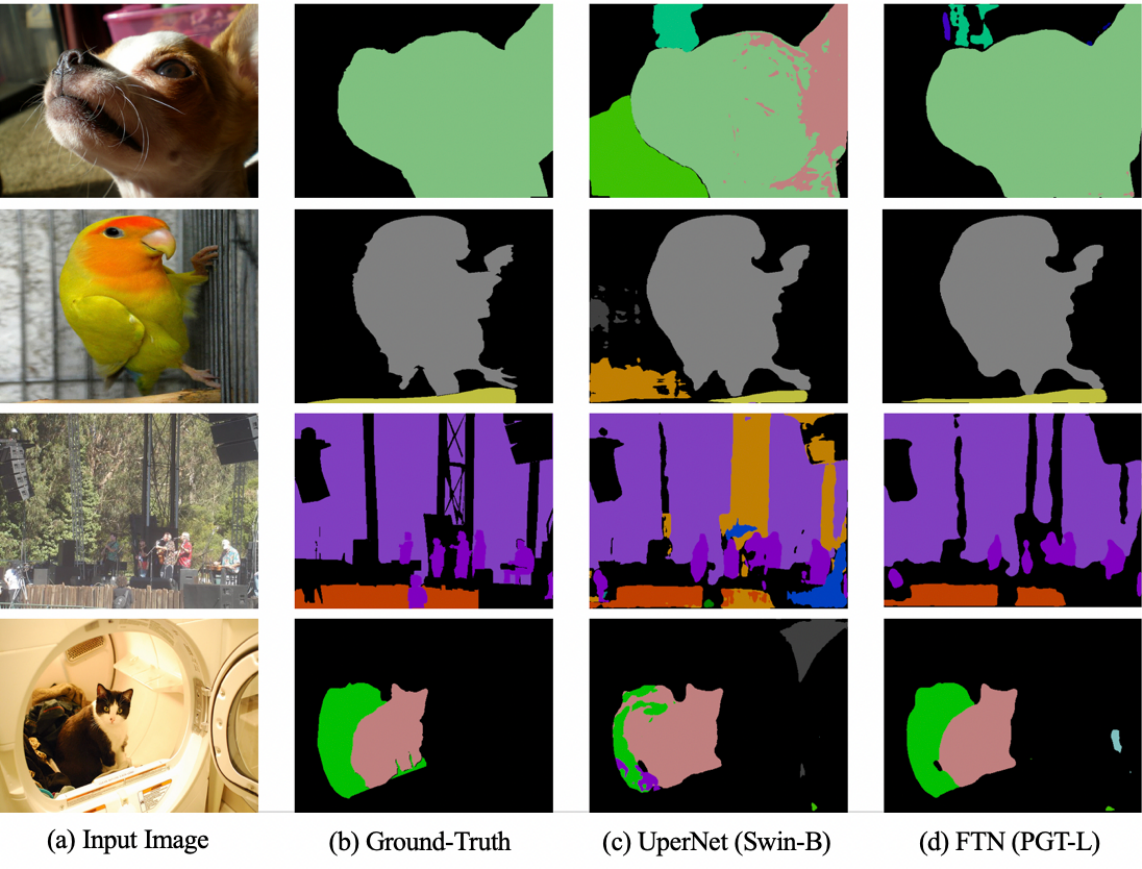}
	\caption{ \label{VIS_PContext} Visualization comparison on PASCAL Context dataset.}
\end{figure}


\noindent 
\textbf{Results on ADE20K.} 
As shown in Table \ref{SEG_EXP}, we compare the performance of FTN with the famous and state-of-the-art models on the more challenging ADE20K validation dataset. Compared with CNN-based models, our FTN has achieved overwhelming advantages with even a lower complexity, such as (48.68\% \emph{vs.} 46.35\%). Compared with other Transformer segmentation models, our FTN models still outperform UperNet(Swin) under the similar computational burden. Besides, our FTN(PGT-B) achieves 50.88\% mIoU, surpassing the more computationally intensive SETR-MLA(ViT-L/16).

\noindent 
\textbf{Results on COCO-Stuff.} 
The state-of-the-art results on COCO-Stuff dataset are shown in Table \ref{SEG_EXP}. RecoNet with ResNet-101 as the backbone achieves a mIoU of 41.50\%, which is the most advanced CNN model. Our FTN(PGT-T) is comparable to the most advanced CNN model (41.57 \emph{vs.} 41.50) with three times reduction in the numbers of parameters. Our FTN(PGT) models are significantly superior to the promising Swin Transformers, with the advantages of about $2 \sim3 \%$. 



\begin{figure}[t]
	\centering
	\includegraphics[width=0.98\linewidth]{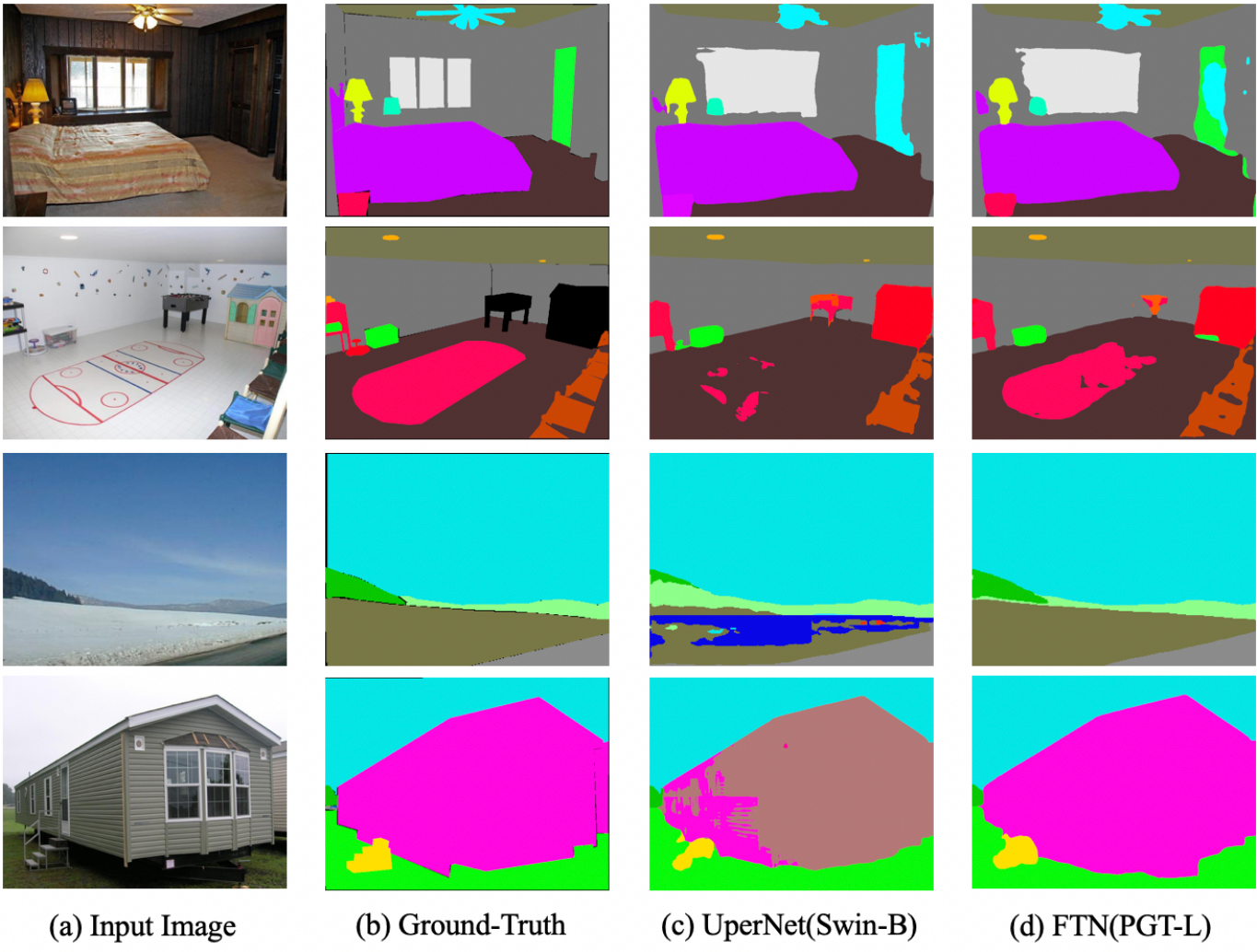}
	\caption{ \label{VIS_ADE20K} Visualization comparison on ADE20K dataset.}
\end{figure}

\begin{figure}[tp]
\centering
\includegraphics[width=0.98\linewidth]{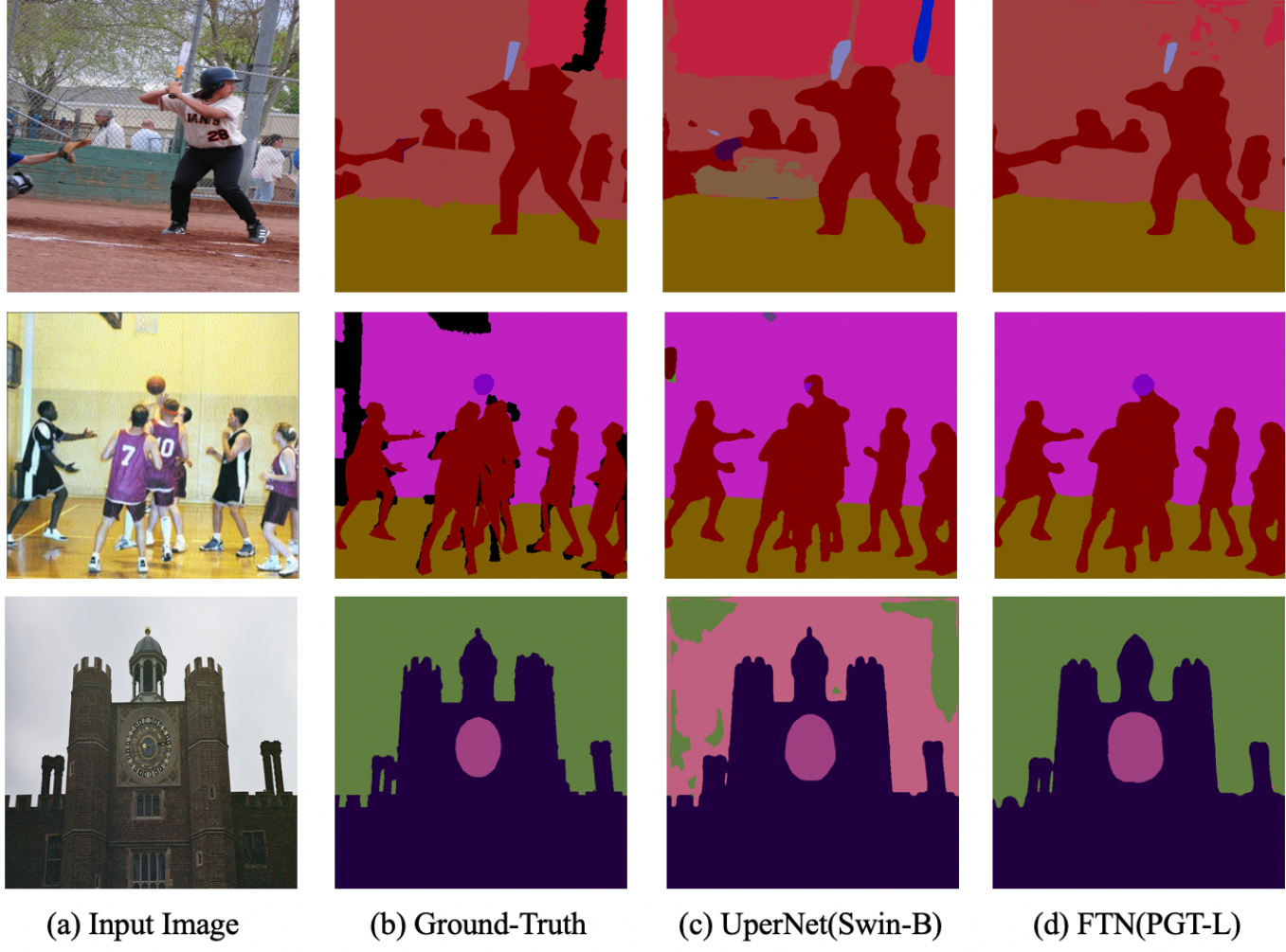}
\caption{ \label{VIS_COCO} Visualization comparison on COCO-Stuff dataset.}
\end{figure}

\noindent 
\textbf{Visualization.} 
We illustrate the visualization results on PASCAL-Context \cite{PContext}, ADE20K \cite{ADE20K} and COCO-Stuff \cite{caesar2018coco} in Figure \ref{VIS_PContext}, Figure \ref{VIS_ADE20K} and Figure \ref{VIS_COCO}, respectively. We mainly compare our FTN with UperNet \cite{upernet}, which performs best in previous Transformer-based segmentation methods. 
According to these visualizations, we can see that our FTN(PGT-L) performs better than UperNet(Swin-B) under similar computational burden in distinguishing the confusing regions with similar colors but different categories. For example, in Line 4 of Figure \ref{VIS_PContext}, UperNet(Swin-B) confused the cat with its surrounding clothing, while our FTN(PGT-L) distinguish them clearly. However, UperNet(Swin-B) will confuse in some visually easy areas. For example, in the second and third row of Figure \ref{VIS_PContext}, the bird and its background are visually distinct, which is visually a simple scene, while UperNet(Swin-B) misjudges some part of the background surprisingly.

\subsection{Face Parsing}
In order to evaluate the generalization of our FTN on other dense prediction tasks, we also apply our FTN(PGT-L) on the face parsing task, which aims to distinguish between different parts of the face, such as eyes, nose, and mouth. 

\noindent 
\textbf{Dataset.} 
We compare our model with the state-of-the-art methods on CelebAMask-HQ dataset \cite{maskgan}, which is a large-scale face parsing dataset with 19 categories including some additional eyeglass, earring, necklace, neck, and cloth beyond the traditional facial components. It contains 24183, 2993, and 2824 images for training, validation, and testing respectively. 

\noindent 
\textbf{Settings.} We conduct experiments on 2 V100 GPUs with a total batch size of 16. The models are trained for 160k iterations with AdamW \cite{AdamW} as the optimizer (weight decay=0.01). The initial learning rate is set to 6e-5 and decay via polynomial scheduler. Synchronized BN and cross-entropy loss are used during training. 

\begin{figure}[t]
\centering
\includegraphics[width=0.98\linewidth]{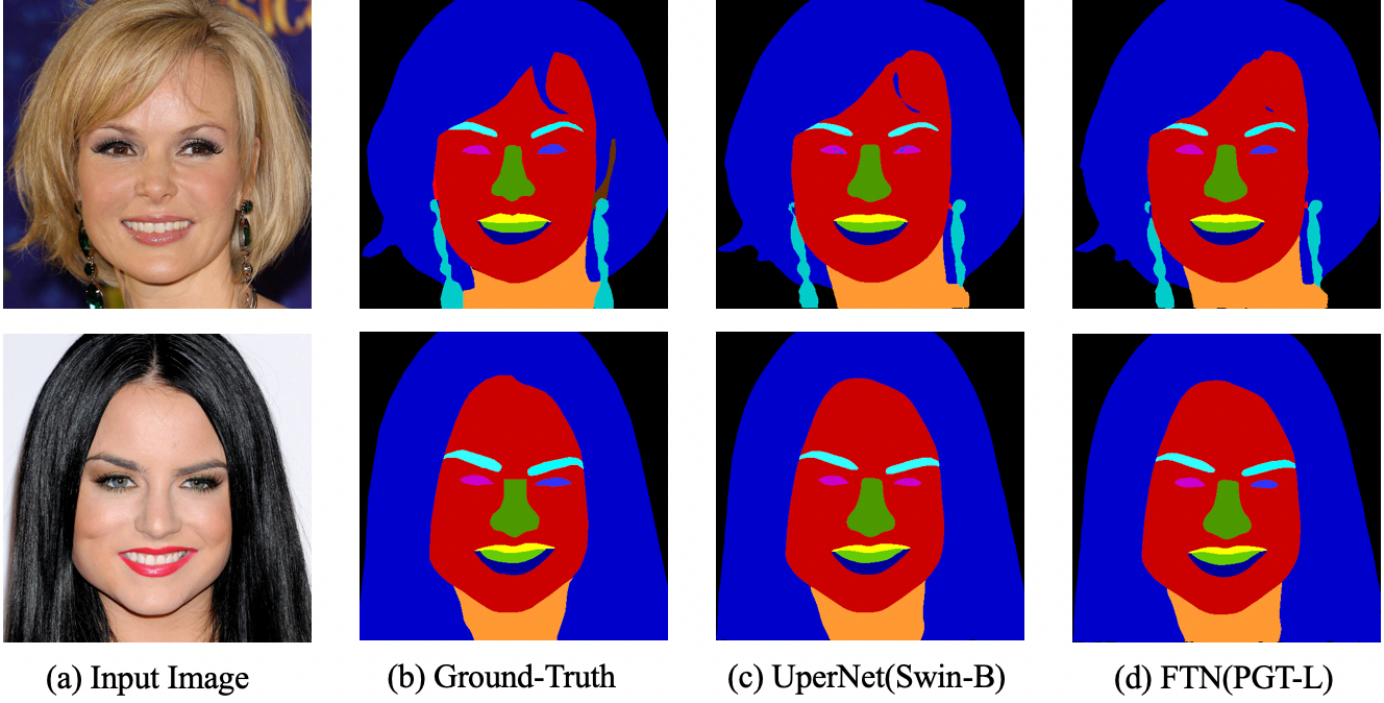}
\caption{ \label{VIS_face} Visualization comparison on CelebAMask-HQ dataset.}
\end{figure}

\noindent 
\textbf{Results.} 
Table \ref{EXP_celeba} compares the F1 score of recent approaches. Compared with previous best CNN-based method, our FTN(PGT-L) outperforms AGRNet \cite{AGRNet} by a large margin (87.4 \emph{vs.} 85.5). Our FTN(PGT-L) also surpasses the UperNet(Swin-B) by +1.2\%. Note that the backbone of both UperNet(Swin-B) and FTN(PGT-L) are pretrained on ImageNet-1k with $224\times224$ input size. Figure \ref{VIS_face} presents the visualization comparisons, which shows that the segmentation of our FTN(PGT-L) is more detailed.

\begin{table*}[tp]
\centering
\caption{\label{EXP_celeba}Comparisons with state-of-the-arts on the CelebAMask-HQ dataset (in F1 score).}
\begin{tabular}{c|ccccccccc|c}
\toprule[1pt]
\multirow{2}{*}{Method}   & Face    & Nose    & Glasses & L-Eye & R-Eye & L-Brow  & R-Brow  & L-Ear  & R-Ear  & \multirow{2}{*}{Mean} \\
& I-Mouth  & U-Lip  & L-Lip  & Hair  & Hat  & Earring & Necklace & Neck  & Cloth   \\
\midrule[1pt]
\multirow{2}{*}{PSPNet \cite{PSPNet}}   & 94.8 & 90.3 & 75.8 & 79.9 & 80.1 & 77.3 & 78.0 & 75.6 & 73.1  & \multirow{2}{*}{76.2} \\
& 89.8 & 87.1 & 88.8 & 90.4 & 58.2 & 65.7 & 19.4 & 82.7 & 64.2 \\
\midrule[0.5pt]
\multirow{2}{*}{MaskGAN \cite{maskgan}}  & 95.5 & 85.6 & 92.9 & 84.3 & 85.2 & 81.4 & 81.2 & 84.9 & 83.1 & \multirow{2}{*}{80.3} \\
& 63.4 & 88.9 & 90.1 & 86.6 & 91.3 & 63.2 & 26.1 & \textbf{92.8} & 68.3 \\
\midrule[0.5pt]
\multirow{2}{*}{EHANet \cite{EHANet}} & 96.0 & 93.7 & 90.6 & 86.2 & 86.5 & 83.2 & 83.1 & 86.5 & 84.1 & \multirow{2}{*}{84.0} \\
& 93.8 & 88.6 & 90.3 & 93.9 & 85.9 & 67.8 & 30.1 & 88.8 & 83.5 \\
\midrule[0.5pt]
\multirow{2}{*}{Wei \textit{et al.} \cite{wei2019accurate}} & 96.4 & 91.9 & 89.5 & 87.1 & 85.0 & 80.8 & 82.5 & 84.1 & 83.3 & \multirow{2}{*}{82.1} \\
& 90.6 & 87.9 & 91.0 & 91.1 & 83.9 & 65.4 & 17.8 & 88.1 & 80.6 \\
\midrule[0.5pt]
\multirow{2}{*}{EAGRNet \cite{EAGRNet}}  & 96.2 & 94.0 & 92.3 & 88.6 & 88.7 & 85.7 & 85.2 & 88.0 & 85.7 & \multirow{2}{*}{85.1} \\
& \textbf{95.0} & 88.9 & 91.2 & 94.9 & 87.6 & 68.3 & 27.6 & 89.4 & 85.3 \\
\midrule[0.5pt]
\multirow{2}{*}{AGRNet \cite{AGRNet}}  & 96.5 & 93.9 & 91.8 & 88.7 & 89.1 & 85.5 & 85.6 & 88.1 & \textbf{88.7} & \multirow{2}{*}{85.5} \\
& 92.0 & 89.1 & 91.1 & 95.2 & 87.2 & 69.6 & 32.8 & 89.9 & 84.9 \\
\midrule[0.5pt]
\multirow{2}{*}{UperNet(Swin-B) \cite{Swin}} & 96.5 & 94.0 & 92.1 & 89.7 & 89.9 & 86.2 & 85.9 & 88.3 & 87.9 & \multirow{2}{*}{86.2} \\
& 92.1 & 89.4 & 91.2 & 95.5 & 90.1 & 70.5 & 35.4 & 90.3 & 87.1  \\
\midrule[0.5pt]
\multirow{2}{*}{FTN-L (ours)} & \textbf{96.6} & \textbf{94.3} & \textbf{93.3} & \textbf{90.0} & \textbf{90.3} & \textbf{87.0} & \textbf{86.6} & \textbf{88.9} & 88.5 & \multirow{2}{*}{\textbf{87.4}} \\
& 92.4 & \textbf{89.9} & \textbf{91.4} & \textbf{96.1} & \textbf{92.2} & \textbf{72.0} & \textbf{40.2} & 92.2 & \textbf{91.0} \\
\bottomrule[1pt]
\end{tabular}
\end{table*}

\subsection{Ablation Study}

In this section, we first replace the encoder and decoder of our FTN with other models for comparisons to demonstrate the effectiveness of PGT and FPT, respectively. Then, we ablate the key designs of PGT, \emph{i.e.}, group numbers, position encoding and class token. Finally, we provide more ablation experiments for some design choices of FPT.


\noindent 
\textbf{Effectiveness of PGT.}
Our proposed FTN is a general and flexible fully Transformer framework for image segmentation, in which the encoder can be replaced by any other Vision Transformer backbones. In order to prove the effectiveness of PGT, we compare the recent promising Transformer backbones, such as ViT \cite{VIT}, PVT \cite{PVT} and Swin Transformer \cite{Swin} with our PGT under various decoders on PASCAL Context. As shown in Table \ref{EFF_EXP}, no matter which decoder is used, our PGT variants are about +3$\sim$6\% higher than PVT variants and +2$\sim$4\% higher than Swin variants on average under similar computation burdens, which demonstrates the superiority of our PGT. It is remarkable that our PGT-L even surpasses the more computationally intensive ViT-L (pretrained on ImageNet-21K) by +1$\sim$2\% improvement with considerably fewer amount of pretraining data.

\noindent 
\textbf{Effectiveness of FPT.}
Meanwhile, the decoder part of FTN is also compatible with various CNN and Transformer-based decoders. To evaluate the effectiveness of our FPT, we compare FPT with the decoders used in previous Transformer segmentation models on PASCAL Context dataset. Experiments in Table \ref{EFF_EXP} show that no matter which backbone is applied, our FPT is consistently superior to other popular decoders, about +1\% higher than Semantic FPN and about +0.5\% higher than UperNet used in \cite{Swin}. Compared with another query-based Transformer decoder proposed in Trans2Seg \cite{Trans2Seg}, our FPT also performs better by a large margin (55.24 \emph{vs.} 54.13).

\begin{table*}[tp]
 \caption{\label{EFF_EXP} Combinations of different encoders and decoders on PASCAL Context validation dataset. $ {\ddagger} $ refers to pretraining on a larger ImageNet-21k dataset under the input size of $384 \times 384$. For a fair comparison, all the settings are same as mentioned in Section \ref{SEG_setup}, except that the auxiliary layers are removed and the total iteration is halved to 40k.}
  \centering
  \scalebox{0.9}{
  \begin{tabular}{c|c|ccccccc}
    \toprule[1pt]
    \multicolumn{2}{c|}{\multirow{2.5}{*}{\textbf{Encoder}}} & \multicolumn{7}{c}{\textbf{Decoder}}      \\ 
    \cmidrule{3-9} 
\multicolumn{2}{c|}{}                         & \multicolumn{5}{c|}{\textbf{CNN}}                                                  & \multicolumn{2}{c}{\textbf{Transformer}} \\ 
\midrule[0.5pt]
    Model                    & Params             & Semantic FPN \cite{semanticfpn} & SETR-PUP \cite{SETR} & SETR-MLA \cite{SETR} & DPT \cite{DPT}  & \multicolumn{1}{c|}{UperNet \cite{upernet}} & Trans2Seg \cite{Trans2Seg}     & FPT(ours)      \\ 
    \midrule[1pt]
    PVT-T \cite{PVT}                  & 13M                & 43.34        & 41.13    & 43.50    & 43.67 & \multicolumn{1}{c|}{43.92}   & 43.03          & 44.62          \\
    PGT-T (ours)                         & 13M                & 49.04        & 46.59    & 49.21    & 48.63 & \multicolumn{1}{c|}{49.95}   & 49.56          & 50.48          \\ 
    \midrule[0.5pt]
    PVT-S \cite{PVT}                 & 25M                & 47.53        & 45.44    & 47.63    & 47.13 & \multicolumn{1}{c|}{47.95}   & 46.19          & 48.43          \\
    Swin-T \cite{Swin}                   & 28M                & 47.01        & 44.43    & 47.29    & 47.25 & \multicolumn{1}{c|}{48.95}   & 47.42          & 49.42          \\
    PGT-S (ours)                         & 28M                & 50.37        & 47.94    & 50.91    & 50.67 & \multicolumn{1}{c|}{51.58}   & 50.83          & 51.93          \\ 
    \midrule[0.5pt]
    Swin-S \cite{Swin}                   & 50M                & 50.03        & 47.10    & 50.77    & 50.51 & \multicolumn{1}{c|}{51.76}   & 50.17          & 52.43          \\
    PVT-L \cite{PVT}                 & 61M                & 49.70        & 47.20    & 50.27    & 49.52 & \multicolumn{1}{c|}{50.53}   & 49.72          & 51.08          \\
    PGT-B (ours)                         & 50M                & 53.26        & 50.29    & 53.73    & 53.24 & \multicolumn{1}{c|}{54.02}   & 53.44          & 54.06          \\ 
    \midrule[0.5pt]
    Swin-B \cite{Swin}                   & 88M                & 50.31        & 47.52    & 51.18    & 52.17 & \multicolumn{1}{c|}{52.51}   & 50.80          & 53.35          \\
    ViT-L/16$^{\ddagger}$ \cite{VIT}    & 307M               & 52.65        & 52.42    & 52.87    & 52.65 & \multicolumn{1}{c|}{52.93}   & 52.78          & 53.43          \\
    PGT-L (ours)                         & 88M                & 54.09        & 52.61    & 54.60    & 54.31 & \multicolumn{1}{c|}{54.73}   & 54.13          & 55.24          \\ 
    \bottomrule[1pt]
  \end{tabular}
  }
\end{table*}

\noindent 
\textbf{Group Choices of PGT.}
As mentioned in Section \ref{PGA}, the receptive fields of PGT can be adjusted by the number of groups in four encoder stages. Excessive receptive fields will impose a heavy computational burden, while if the receptive field is too small, it will be insufficient for context modeling. Figure \ref{Ablation_1} shows the comparison with different choices of PGT group numbers in four encoder stages. PGT$<$64-16-1-1$>$ is +0.38\% better than PGT$<$64-16-4-1$>$ with fewer additional computation, which demonstrates that global context modeling on the 1/16 resolution features in stage3 is critical to the performance. PGT$<$16-4-1-1$>$ is slightly +0.02\% better than PGT$<$64-16-1-1$>$, but much more expensive for memory cost. As a result, we use PGT$<$64-16-1-1$>$ in our PGT variants.

\noindent 
\textbf{Position Encoding of PGT.}
Positional embeddings are used to introduce the awareness of spatial location for self-attention, which have been proved to be critical to Transformers \cite{Transformer, VIT, Swin, CPVT}. To understand the effect of position encoding methods for our PGT, we compare four different settings, namely no position encoding (no pos.), learnable absolute position encoding (APE) \cite{VIT}, conditional position encoding (CPE) \cite{CPVT} and a combination of APE and CPE (APE\&CPE). As shown in Figure \ref{Ablation_pos}, APE achieves +1.16\% higher than no pos., which shows the importance of position encoding. CPE achieves +0.7\% higher than APE and +0.34\% higher than APE\&CPE. Furthermore, CPE is compatible with variable-length input sequences, which is an obvious limitation of absolute position encoding. Therefore, we use CPE in our PGT as an implicit position encoding scheme.

\begin{table}[t]
\centering
\caption{\label{Ablation_1}Different receptive fields of PGT. The a, b, c, d in $<$a-b-c-d$>$ represent the group number in stage 1$\sim$4 respectively.}
\scalebox{1.0}{
\begin{tabular}{p{3.5cm}|p{3.5cm}}
\toprule[1pt]
\makecell[c]{Pyramid Group Numbers}          & \makecell[c]{Top-1 Acc. (\%)} \\ 
\midrule[1pt]
\makecell[c]{$<$64-16-4-1$>$}         & \makecell[c]{77.81}    \\
\makecell[c]{$<$64-16-1-1$>$}         & \makecell[c]{78.19}    \\
\makecell[c]{$<$16-4-1-1$>$}           & \makecell[c]{\textbf{78.21}}    \\
\bottomrule[1pt]
\end{tabular}}
\end{table}

\begin{table}[t]
\centering
\caption{\label{Ablation_pos}Different position encoding approaches in PGT.}
\scalebox{1.0}{
\begin{tabular}{p{3.5cm}|p{3.5cm}}
\toprule[1pt]
\makecell[c]{Position Encoding}          & \makecell[c]{Top-1 Acc. (\%)} \\ 
\midrule[1pt]
\makecell[c]{no pos.}              & \makecell[c]{76.33}    \\
\makecell[c]{APE}                  & \makecell[c]{77.49}    \\
\makecell[c]{CPE}                  & \makecell[c]{\textbf{78.19}}    \\
\makecell[c]{APE\&CPE}         & \makecell[c]{77.85}    \\
\bottomrule[1pt]
\end{tabular}}
\end{table}

\noindent  
\textbf{CLS-Token of PGT.}
CLS-Token plays the role of a global image representation, which is expected to contain all of discriminate information for final classification. Some of Vision Transformers kept CLS-Token throughout the network \cite{VIT}, while some works only involved it at the last stage \cite{PVT}. Considering that CLS-Token may have missed some key information, \cite{Swin} directly removed the CLS-Token and applied average pooling on the last output of the backbone instead. Inspired by above all, we study the impact of four CLS-token patterns for PGT. As shown in Figure \ref{Ablation_clstoken}, average pooling performs the best, achieving +0.37\% and +0.48\% higher than the CLS-token in only stage4 and all stages respectively. Thus, in PGT, we simply apply average pooling on the final output of stage 4 to generate a global token rather than CLS-Token.

\begin{table}[t]
\centering
\caption{\label{Ablation_clstoken}Different CLS-Token in PGT.}
\scalebox{1.0}{
\begin{tabular}{p{3.5cm}|p{3.5cm}}
\toprule[1pt]
\makecell[c]{CLS-Token}            & \makecell[c]{Top-1 Acc. (\%)} \\ 
\midrule[1pt]
\makecell[c]{stage 4}                  & \makecell[c]{77.82}    \\
\makecell[c]{stage 1$\sim$4}               & \makecell[c]{77.71}    \\
\makecell[c]{avgpooling}             & \makecell[c]{\textbf{78.19}}    \\
\bottomrule[1pt]
\end{tabular}}
\end{table}

\begin{table}[t]
\centering
\caption{\label{Ablation_FPT}Comparisons of different settings of FPT. $<$a-b-c$>$ represents the settings for the transformer blocks which are performed on the feature maps with stride 32("a"), 16("b") and 8("c"), respectively. The mIoU is obtained by multi-scale inference on COCO-Stuff validation dataset.}
\scalebox{0.9}{
\begin{tabular}{cccc|c}
\toprule[1pt]
Depth                          & Embedding dim       & SR-MSA ratio         & Fusion mode        & mIoU (\%) \\ 
\midrule[1pt]
\textless{}1-1-1\textgreater{} & 256           & \textless{}2-2-2\textgreater{}   & add           & 42.41   \\
\textless{}1-1-1\textgreater{} & 512           & \textless{}2-2-2\textgreater{}   & add           & \textbf{43.37} \\
\textless{}1-2-1\textgreater{} & 512           & \textless{}2-2-2\textgreater{}   & add           & 43.11   \\
\textless{}1-1-1\textgreater{} & 512           & \textless{}1-2-2\textgreater{}    & add           & 43.26   \\
\textless{}1-1-1\textgreater{} & 512           &  \textless{}1-1-2\textgreater{}    & add           & 43.33 \\
\textless{}1-1-1\textgreater{} & 512           & \textless{}2-2-2\textgreater{}    & concat       & 43.35   \\ 
\bottomrule[1pt]
\end{tabular}}
\end{table}

\noindent  
\textbf{Scales of FPT.}
With the limited segmentation training data, the scale of decoder is not the larger the better for segmentation \cite{Trans2Seg}, so that there is an important trade-off between the scale of FPT and the performance. In our FPT, the scale is mainly determined by depth, embedding dim, and the reduction ratio of SR-MSA. We compare several combinations of these hyper-parameters with PGT-S as the backbone on COCO-Stuff validation set. All the settings are the same as Section 4.2, except the abandoned auxiliary layers.

Considering that features with stride 16 have a good trade-off in terms of containing semantic and detailed information, we try to enlarge the depth (number of blocks) for Transformer blocks with 1/16 features in FPT. As shown in Table \ref{Ablation_FPT}, increasing the depth for 1/16 features results in a 0.26\% mIoU drop (43.37 \emph{vs.} 43.11), which indicates that the FPT decoder is not the deeper the better.

Embedding dim refers to the dimension of tokens in transformer blocks of FPT. Excessive small embedding dim might lead to the lack of representation ability, and a too large one may result in channel redundancy and increased computational burden. Considering the constraints of computing resources, we tried two computation-friendly choices of embedding dims (\emph{dim=256} and \emph{512}). As shown in Table \ref{Ablation_FPT}, \emph{dim=512} is significantly superior to \emph{dim=256} (43.37 \emph{vs.} 42.41). Therefore, we set the embedding dim to 512 in our FPT.

SR-MSA is used to reduce memory and computation cost by spatially reducing the number of key and value tokens, especially for high-resolution representations. Considering that the token numbers of the low-resolution features are relatively fewer, we attempt to remove the spatial reduction for transformer blocks with 1/32 and 1/16 features. Table \ref{Ablation_FPT} compares the different choices of reduction ratio of SR-MSA in FPT. It turns out that without using SR-MSA on low-resolution representations, the computation cost is increased, while there is no significant performance gain as expected (43.37 \emph{vs.} 43.33 \emph{vs.} 43.26). On the contrary, not using SR-MSA results in slight performance degradation.

\noindent  
\textbf{Fusion Mode of FPT.}
In our FPT, the multi-level high-resolution features of each branch need to be fused to generate finer prediction. We explore two simple fusion mode, \emph{i.e.}, element-wise summation and channel-wise concatenation. As shown in Table \ref{Ablation_FPT}, summation performs +0.02\% slightly better than concatenation, thus we fuse the multi-level representations by the simpler and more efficient element-wise summation.


\section{Conclusion}
We have developed the Fully Transformer Networks (FTN) for semantic image segmentation. The core contributions of FTN are the proposed Pyramid Group Transformer (PGT) encoder and Feature Pyramid Transformer (FPT) decoder. The former learns hierarchical representations, which creates Pyramid Group Multi-head Self Attention to dramatically improve the efficiency and reduce the computation costs of the traditional Multi-head Self Attention. The latter fuses semantic-level and spatial-level information from multiple stages of the encoder, which can promotes to the generation of finer segmentation results. Extensive experimental results on PASCAL-Context, ADE20K, and COCO-Stuff have shown that our FTN can outperform the state-of-the-art methods in semantic image segmentation, demonstrating that the Fully Transformer Networks can achieve better results than hybrid Transformer and CNN approaches.

{\small
\bibliographystyle{ieee_fullname}
\bibliography{egbib}
}

\end{document}